\documentclass[a4paper]{article}

\usepackage[dvipsnames]{xcolor}

\definecolor{c1}{HTML}{a1c9f4}
\definecolor{c2}{HTML}{ffb482}
\definecolor{c3}{HTML}{8de5a1}
\definecolor{c4}{HTML}{ff9f9b}

\usepackage{graphicx}
\usepackage{hyperref}
\usepackage{verbatim}

\usepackage{latexsym,amsmath,amssymb,amsthm,amscd} 

\usepackage{booktabs}
\usepackage{todonotes}
\usepackage[capitalise]{cleveref}
\usepackage{subcaption}

\usepackage{mathtools}
\newcommand{\T}{\mathcal{T}}

\newcommand{\D}{\mathcal{D}}

\usepackage{tikz}
\usetikzlibrary{positioning,arrows.meta,shapes}

\begin{document}
\title{Continuous Fair SMOTE -- Fairness-Aware Stream Learning from Imbalanced Data}
\author{Kathrin Lammers \and
Valerie Vaquet \and
Barbara Hammer}

\date{Machine Learning Group \\
Bielefeld University, Bielefeld - Germany}

\maketitle

\begin{abstract}
As machine learning is increasingly applied in an online fashion to deal with evolving data streams, the fairness of these algorithms is a matter of growing ethical and legal concern. In many use cases, class imbalance in the data also needs to be dealt with to ensure predictive performance. Current fairness-aware stream learners typically attempt to solve these issues through in- or post-processing by focusing on optimizing one specific discrimination metric, addressing class imbalance in a separate processing step. While C-SMOTE is a highly effective model-agnostic pre-processing approach to mitigate class imbalance, as a side effect of this method, algorithmic bias is often introduced.

Therefore, we propose CFSMOTE - a fairness-aware, continuous SMOTE variant - as a pre-processing approach to simultaneously address the class imbalance and fairness concerns by employing situation testing and balancing fairness-relevant groups during oversampling. Unlike other fairness-aware stream learners, CFSMOTE is not optimizing for only one specific fairness metric, therefore avoiding potentially problematic trade-offs. Our experiments show significant improvement on several common group fairness metrics in comparison to vanilla C-SMOTE while maintaining competitive performance, also in comparison to other fairness-aware algorithms. 

\end{abstract}

\section{Introduction}

\footnotetext{Funding in the scope of the BMBF project KI Akademie OWL under grant agreement No 01IS24057A is gratefully acknowledged.}

We increasingly rely on machine learning algorithms for many different aspects of life, such as communications, healthcare, infrastructure, education, and finance. The question of whether these algorithms are \textit{fair} is of ever-increasing importance to society \cite{review_fairness_ml}. Legislation, such as the EU AI Act, demands that algorithmic decision-making is free from discrimination and unfair biases \cite{ai_act_2024}, and a growing body of work has emerged in the area of fair machine learning in recent years \cite{review_fairness_ml}.

Simultaneously, \textit{stream learning} has been an area of growing interest in a world where real-time applications in areas such as e-commerce or stock trading operate not on static data but on data streams, where individual samples arrive in real-time and need to be processed immediately. Additional challenges in this setting include \emph{concept drift}, a change in the underlying data distribution of the stream, to which the classifier needs to adapt swiftly to retain performance \cite{concept_drift_survey}. More widely known machine learning issues such as \emph{class imbalance} also require addressing, particularly if class distributions change as a result of drift \cite{hoi_online_2021}.

While fairness has become a well-explored topic in the context of \textit{batch learning}, fairness concerns in the context of \textit{stream learning} are still largely under-explored. In recent years, several fair stream learning algorithms, such as Fairness-Aware Hoeffding trees (FAHT) \cite{FAHT} or FABBOO \cite{FABBOO}, have been proposed. However, these algorithms typically define fairness in terms of statistical parity of outcomes between groups. Other aspects of fairness - such as similar treatment of similar individuals - are usually not addressed. Instead, post-processing approaches, in particular, might deliberately treat otherwise identical individuals differently based on their sensitive attributes to ensure group fairness. Thereby, concerns of individual fairness get largely ignored. This is a research gap which we aim to address in this work. 

In the static setting, FairSMOTE \cite{fairsmote} has been proposed as a pre-processing technique to directly increase individual fairness, while also promoting parity between groups to faithfully reflect the data stream. Experiments have shown that it reduces statistical parity between groups in comparison to conventional SMOTE while simultaneously promoting similar treatment for similar individuals through situation testing.

Although a continuous version of SMOTE exists and has been proven to be useful for addressing class imbalance in the stream learning setting \cite{csmote}, it has not yet been adapted to address issues of fairness. Instead, comparison with existing fair stream learners, such as FABBOO, has shown that C-SMOTE exacerbates discrimination by producing results that show significantly greater disparity of outcomes between groups than inherent to the data \cite{FABBOO}.

To address these issues, we propose CFSMOTE, a fairness-aware, continuous SMOTE variant which adapts the advantages of FairSMOTE to the stream learning setting. As a pre-processing method, CFSMOTE can be combined with any base classifier and enhances fairness simply by oversampling minority data points without the need to explicitly state fairness as an objective or to know which group is discriminated against and to what extent.

In this paper, we will first outline relevant problem definitions, both for stream learning as well as for fairness in \cref{sec:foundations}, and then give a brief overview of the relevant related work (\cref{sec:related_work}) before presenting an in-depth description of the CFSMOTE algorithm in \cref{sec:cfsmote}. Finally, we evaluate our algorithm and compare its performance to both vanilla C-SMOTE and available competitors on a number of performance and fairness metrics through experiments (\cref{sec:experiments}) on relevant fairness datasets. \Cref{sec:conclusion} concludes this work.

\section{Foundations \label{sec:foundations}} 

\subsection{Learning from Data Streams}
Machine learning is usually practised in the batch set-up, where the entire training dataset is fully available throughout the training process preceding evaluation and deployment. In contrast, in \emph{stream learning} for classification, we consider a data stream $\Sigma={(x_t, y_t)}_{t=1}^T$ with $x_t\in X$ and $y_t\in Y $ generated by a time-indexed distribution family $(\D_t)_{t \in \T}$ with time domain $\T$. Training and evaluation are performed in the \emph{test-then-train} set-up: Whenever a new sample $x_t$ becomes available, it is first used for evaluating the performance of the classifier. Afterward, potentially with some delay, the true label $y_t$ becomes available and can be used to update the classifier. This way, an independent performance estimate can be obtained \cite{concept_drift_survey}.

One of the main challenges in online learning is distributional changes, commonly referred to as \emph{concept drift} or drift for shorthand. Drift occurs if $\D_{t_0}\neq \D_{t_1}$ for at least two time points $t_0, t_1$. To keep its predictive performance in such settings, the online learner must quickly adapt to the change \cite{concept_drift_survey}. Additionally, \emph{class imbalance} presents a challenge in stream learning similar to batch learning, with the additional complication that the class distribution may also be affected by concept drift and change over time. In the case of imbalanced data, alternative metrics not neglecting minority class performance, such as the \textit{balanced accuracy}, are required for evaluation.

\subsection{Fairness in Machine Learning}

Fairness is a topic of increasing relevance in machine learning, with many different definitions formalizing fairness emerging \cite{review_fairness_ml}. Most fairness notions can be either categorized into group or individual fairness. The first focuses on similar treatment for distinct groups, while the latter focuses on a similar treatment of similar individuals. In this section, we will briefly define the most popular definitions. Let $\hat{y_t}$ be the predicted and $y_t$ the true label, with $1$ indicating positive outcomes (w.l.o.g.). Let $S$ be the sensitive attribute, with $S = 1$ representing the privileged group, while $A$ represents all other attributes. The $\epsilon$ represents the greatest acceptable, measured level of \textit{discrimination}; therefore, smaller is better for these metrics.

Popular group fairness definitions aim at providing the same possibilities to the considered groups by enforcing similarity with respect to positive outcomes. \emph{Disparate impact} aims at a similar rate of positive outcomes $ 1 - \frac{P [ \hat{y_t}=1 | S \neq 1]}{P [ \hat{y_t}=1 | S = 1]} \leq \epsilon$ while \emph{demographic parity} (also frequently referred to as \textit{statistical parity}) considers absolute difference rather than fractions: $| P [ \hat{y_t}=1 | S = 1] - P [ \hat{y_t}=1 | S \neq 1] | \leq \epsilon $. As they ignore possible disparities inherent to the data, relying on them creates a \emph{fairness-accuracy-trade-off} but enforces fairness even in inherently biased datasets.

Alternatively, assuming that the data and the ground truth labels are fair, one can rely on penalizing certain types of errors in the classification: While \emph{equal opportunity} (\textit{equal false-negative rates}) states that for samples with a positive label, the rates of positive prediction should be similar, i.e. $| P [ \hat{y_t}=1 | S = 1, y_t = 1] - P [ \hat{y_t}=1 | S \neq 1, y_t = 1] | \leq \epsilon $, in \emph{predictive equality} (\textit{equal false-positive rates}), the rates of positive predictions for samples with negative ground truth labels should be similar, i.e. $| P [ \hat{y_t}=1 | S = 1, y_t = 0] - P [ \hat{y_t}=1 | S \neq 1, y_t = 0] | \leq \epsilon$. \emph{Equalized odds} enforces both at the same time.

Finally, the most direct notion of individual fairness is \emph{fairness through unawareness}, which states  that the sensitive attribute label should not affect predictive outcome:
$|P(\hat{y_t} = y_t | S, A) - P(\hat{y} = y_t | S^{-1}, A)| \leq \epsilon$. This definition is simple but often not enough to account for possible issues with proxy attributes.

Fairness-enhancing strategies can be categorized into \emph{pre-, in-, and post-processing strategies}. While pre-processing methods try to enhance the inherent fairness in the data, in-processing approaches adapt the training procedure of a model, and post-processing adapts the model's decisions after training. We will present different methodologies for implementing these strategies in the next section.

\section{Related Work \label{sec:related_work}}
Learning from imbalanced data streams has received growing interest in the last few years \cite{aguiar_survey_2024}. Many novel methodologies have been proposed, with the majority implementing some pre-processing strategy, frequently relying on resampling schemes. 
For instance, C-SMOTE \cite{csmote} adapts the popular Synthetic Minority Oversampling Technique (SMOTE) \cite{chawla_smote_2002} to online learning through a window-based approach, keeping windows of both classes of samples as the basis for synthetic up-sampling and utilizing ADWIN \cite{bifet_learning_2007} for drift detection.

Recently, fair online learning has become an area of interest, and some fair online classifiers have been proposed. Many works implement in-processing techniques relying on fair Heoffding tree variants, e.g. FAHT \cite{FAHT}, FEAT \cite{feat}, and 2FAHT \cite{2FAHT}. For instance, FAHT has a modified split metric that takes fairness in terms of statistical parity into account.
FABBOO \cite{FABBOO}, meanwhile, is an imbalance-aware boosting approach which ensures fairness by adjusting its decision boundary in a post-processing step to optimize either demographic parity or equal opportunity with the tolerated discrimination as a hyper-parameter.

In the batch learning setting, SMOTE has been adapted to address the issue of fairness. FairSMOTE \cite{fairsmote} accomplishes this by dividing the data points into four groups - based on both class labels as well as sensitive attribute values - and ensures that each group is represented by a given minimum percentage of samples. For all groups where the original number of samples is insufficient, synthetic samples are generated based on original samples from the group. 
In addition to upsampling minority groups, FairSMOTE also performs \emph{situation testing} before any sample - original or synthetic - is used to train the actual base classifier. For this purpose, a simple proxy model - a logistic regression classifier - is first trained on and subsequently used to predict all data points, both original and synthetic. Additionally, this proxy model is also used to predict counterfactuals of each data point where the sensitive attribute value is flipped. If a data point and its corresponding counterfactual receive different class labels, that data point gets discarded and is not used to train the actual base learner.

\section{CFSMOTE \label{sec:cfsmote}}

To transfer the fairness-enhancing properties of FairSMOTE to the stream learning setting, we propose CFSMOTE, short for Continuous Fairness-aware Minority Oversampling Technique.

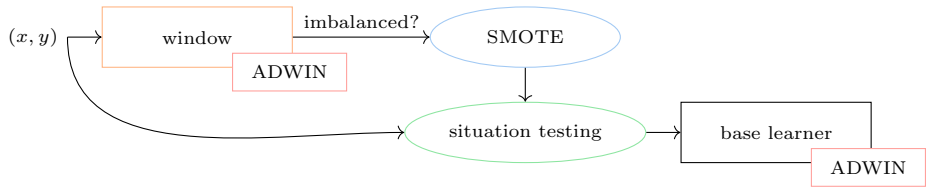
\begin{figure}[t]    \centering
    \begin{tikzpicture}[node distance=0.45cm]
    \tikzstyle{every node}=[font=\scriptsize]
\node (sample) {$(x,y)$};
\node [right = of sample](window) [draw,c2,minimum height=0.8cm, minimum width=2.5cm] {\textcolor{black}{window}};
\node [ right = 1.8cm of window] (smote) [draw, c1,minimum height=0.8cm, minimum width=2.5cm, ellipse] {\textcolor{black}{SMOTE}}; 
\node [below = of smote] (situation) [draw, c3, minimum height=0.8cm, minimum width=2.5cm, ellipse]  {\textcolor{black}{situation testing}};
\node [right = of situation] (learner) [draw, minimum height=0.8cm, minimum width=2.5cm] {base learner};

\node [below right = -0.2cm and -0.8cm of learner] [draw, c4, fill=white, minimum height=0.5cm, minimum width=1.5cm] {\textcolor{black}{ADWIN}};
\node [below right = -0.2cm and -0.8cm of window] [draw, c4, fill=white, minimum height=0.5cm, minimum width=1.5cm] {\textcolor{black}{ADWIN}};
\draw[->] (sample.east) to (window.west);
\draw[->] (sample.east) to [out=270,in=180] (situation.west);
\draw[->] (window.east) to node[above] {imbalanced?} (smote.west);
\draw[->] (smote.south) to (situation.north);
\draw[->] (situation.east) -- (learner.west);

\end{tikzpicture}
    \caption{An overview of the CFSMOTE algorithm. A new sample must pass situation testing before it gets passed to the base learner. Simultaneously, it is used to update the window of current samples. If imbalance is detected, SMOTE-based oversampling for the minority group is triggered. Synthetic samples do not get stored, but the imbalance rates get updated after generation, and the samples get passed on to situation testing. The window of samples is constantly monitored by an ADWIN drift detector and adjusted accordingly.}
    \label{fig:flowchart_cfsmo}
  \end{figure}

As visualized in \cref{fig:flowchart_cfsmo}, similar to CSMOTE, CFSMOTE follows a \emph{window-based approach} \textcolor{c2}{\textbullet}, constantly updating windows with current samples from the stream which are later used for creating synthetic samples. Unlike C-SMOTE, we do not only divide our window of samples according to labels but split those two groups again based on the sensitive attribute. That way, we end up with \emph{four} sub-groups: Unprivileged and negative samples, 
unprivileged and positive samples, 
privileged and negative samples, 
and privileged and positive samples. 
Note that whether a specific sensitive attribute value is described as privileged or unprivileged is irrelevant to our classifier, as all four groups are essentially treated equally.

As in C-SMOTE, the window of samples also gets constantly monitored for drift, with dedicated ADWIN drift detectors  \textcolor{c4}{\textbullet} monitoring changes in the distribution of a) the class labels and b) the sensitive attribute values. As long as no drift is detected, the window size increases with every new sample, ensuring a high level of diverse samples for oversampling. If drift is detected, however, the last detected drift point is used to determine the new window size for the oversampling process, while older samples get discarded and the imbalance ratio adjusted to disregard synthetic samples generated based on outdated original data points. 

To adapt to class imbalance and an under-representation of a group, an imbalance check is performed \textcolor{c1}{\textbullet}.
The imbalance ratio is calculated based on the sum of both natural and synthetic samples in each sub-group of the current window. 

When the \emph{minimum imbalance ratio}, a pre-defined threshold, is undercut and all sub-groups meet the \emph{minimum size} necessary for oversampling, synthetic samples are generated in typical SMOTE fashion \cite{chawla_smote_2002} based on the original samples of the minority group(s) in the current window until the minimum imbalance ratio is met.

Finally, before either natural or synthetic samples are used to train the base learner, each sample undergoes \textit{situation testing}  \textcolor{c3}{\textbullet}, where the prediction for the actual sample and its equivalent with a switched sensitive attribute label are compared. If the predictions are not the same, the sample is assumed to promote individual unfairness and, therefore, not used for training the base learner. 

The base learner is also monitored by an ADWIN drift detector and gets reset if its error increases in a manner that indicates drift. As a pre-processing approach, CFSMOTE is compatible with any stream learning classifier.

\section{Experiments \label{sec:experiments}}
In our experiments, we will first compare the performance of CFSMOTE and C-SMOTE both in terms of fairness as well as performance. Additionally, we will compare the performance of CFSMOTE with other fairness-aware stream learning algorithms, namely the imbalance-aware FABBOO and FAHT, the latter of which we use as a baseline which is not equipped to deal with class imbalance.
\subsection{Datasets}
Although the importance of fairness in machine learning research has been continuously increasing in recent years, there is a lack of suitable benchmark datasets in the area, particularly as it concerns stream learning. Due to the lack of non-stationary datasets that are characterised both by class imbalance as well as sufficient inherent demographic disparity, we chose to evaluate our algorithms on two of the most commonly used stationary fairness benchmark datasets, \textit{Adult}\cite{adult_dataset} and \textit{KDD Census} \cite{noauthor_census-income_2000}. Both datasets are binary classification datasets for the task of income prediction and have \textit{sex} as the sensitive attribute.
To convert these datasets into streams, we repeatedly randomly permuted the order of samples to generate a total of ten data streams from each dataset. Characteristics of the datasets can be found in \cref{tab:datasets}. We did not consider the popular \textit{Compas} \cite{mattu_machine_nodate} dataset as it does not exhibit a substantial degree of class imbalance.

\begin{table}[t]

\caption{Dataset Characteristics}
\label{tab:datasets}
\centering
\scriptsize

\setlength{\tabcolsep}{0.4em}
\begin{tabular}{lrccccccc}
\toprule
Dataset & \#Samples & \#Attr. & $S$ & $S = 0$ & $S = 1$ & Imb. (+\%) & Disp. Impact & Stat. Parity\\
\midrule
Adult & 48842 & 14 & \textit{sex} & \textit{F} & \textit{M} & 23.93\% & 64.03\% & 19.45\%\\
KDD Census & 299285 & 40 & \textit{sex} & \textit{F} & \textit{M} & 6.20\% & 74.96\% & 7.63\% \\
\bottomrule
\end{tabular}
\end{table}

\subsection{Setup}
For our comparison with the C-SMOTE algorithm, we used a Hoeffding Adaptive Tree \cite{HoeffdingAdaptiveTrees} as base learner, while for the comparison with other fairness-aware stream learning algorithms, an Adaptive Random Forest \cite{AdaptiveRandomForests} was chosen as the base classifier to ensure a fair comparison with the ensemble method FABBOO. 
In the latter case, the ensemble size was set to 10 for all classifiers. For both CFSMOTE and C-SMOTE, we set $k=3$ for the kNN algorithm employed during the upsampling stage and used a \textit{minimum window size} of 10. For FABBOO, we optimize separately for statistical parity (FABBOO SP) and equal opportunity (EQOP).
The \textit{minimum imbalance ratio} was set as the default value of $0.5$ for C-SMOTE and $0.245$ for CFSMOTE. Our CFSMOTE implementation\footnote{To ensure anonymity, the code will be made available upon acceptance.} is compatible with the Python \textit{river} stream learning framework \cite{river_library}.

\subsection{CFSMOTE vs CSMOTE}
\begin{table}[t]
\caption{Comparison of CSMOTE and CFSMOTE\label{tab:cf-vs-csmote}}
\scriptsize
\centering

\setlength{\tabcolsep}{0.4em}
\begin{tabular}{l|ll|lll}
    \toprule
      & \multicolumn{2}{c}{Adult} & \multicolumn{2}{|c}{KDD}  \\
     & \multicolumn{1}{c}{CFSMOTE} 
    & \multicolumn{1}{c}{CSMOTE}      & \multicolumn{1}{|c}{CFSMOTE} 
    & \multicolumn{1}{c}{CSMOTE}  \\
    \midrule
    Accuracy & $77.24 \pm 1.10 \; (- 0.28\%)$& $77.46 \pm 1.05$ & $79.72 \pm 0.82 \; (- 5.44 \%)$ & $84.31 \pm 0.99$ \\
    Balanced Accuracy & $76.09 \pm 0.91 \;  (- 2.90\%)$& $78.36 \pm 0.53$ & $80.10 \pm 0.67 \; (+ 1.61 \%)$ & $78.83 \pm 2.36$ \\
    Recall & $73.90 \pm 2.45 \;  (- 7.73\%)$ & $80.09 \pm 0.94$ & $80.53 \pm 0.68 \; (+ 10.94 \%)$ & $72.56 \pm 6.15$ \\
    Geometric Mean & $76.04 \pm 0.94 \;  (- 2.92 \%)$ & $78.33 \pm 0.55$ & $80.09 \pm 0.67 \; (+ 2.06 \%)$ & $78.47 \pm 2.91$ \\
    \midrule
    Statistical Parity & $20.17 \pm 3.76 \;  (- 34.59  \%)$ & $30.84 \pm 1.52$& $\:\;7.70 \pm 0.47 \; (- 34.63\%)$ & $11.78 \pm 0.20$ \\
    Disparate Impact & $48.96 \pm 5.84 \;  (- 22.93  \%)$ & $63.53 \pm 1.98$ & $27.46 \pm 1.98 \; (- 38.00\%)$ & $44.29 \pm 1.08$ \\
    Equal Opportunity & $\:\;3.04 \pm 1.71 \;  (- 75.10  \%)$ & $12.21 \pm 1.98$ & $\:\;1.27 \pm 0.44 \; (- 85.92\%)$ & $\:\;9.02 \pm 0.71$ \\
    Equal FPR & $12.44 \pm 4.00 \;  (- 44.37  \%)$ & $22.32 \pm 1.67$ & $\:\;3.28 \pm 0.40 \; (- 53.67\%)$ & $\:\;7.08 \pm 0.23$ \\
    Individual Fairness & $\:\;0.83 \pm 0.34 \;  (- 27.19  \%)$ & $\:\;1.14 \pm 0.17$ & $\:\;0.36 \pm 0.05 \; (- 76.00\%)$ & $\:\;1.50 \pm 0.08$ \\
    \bottomrule
    \end{tabular}
\end{table}

\begin{figure}[t]
\centering
\centering
     \begin{subfigure}[b]{0.5\textwidth}
         \centering      
         \includegraphics[width=\linewidth]{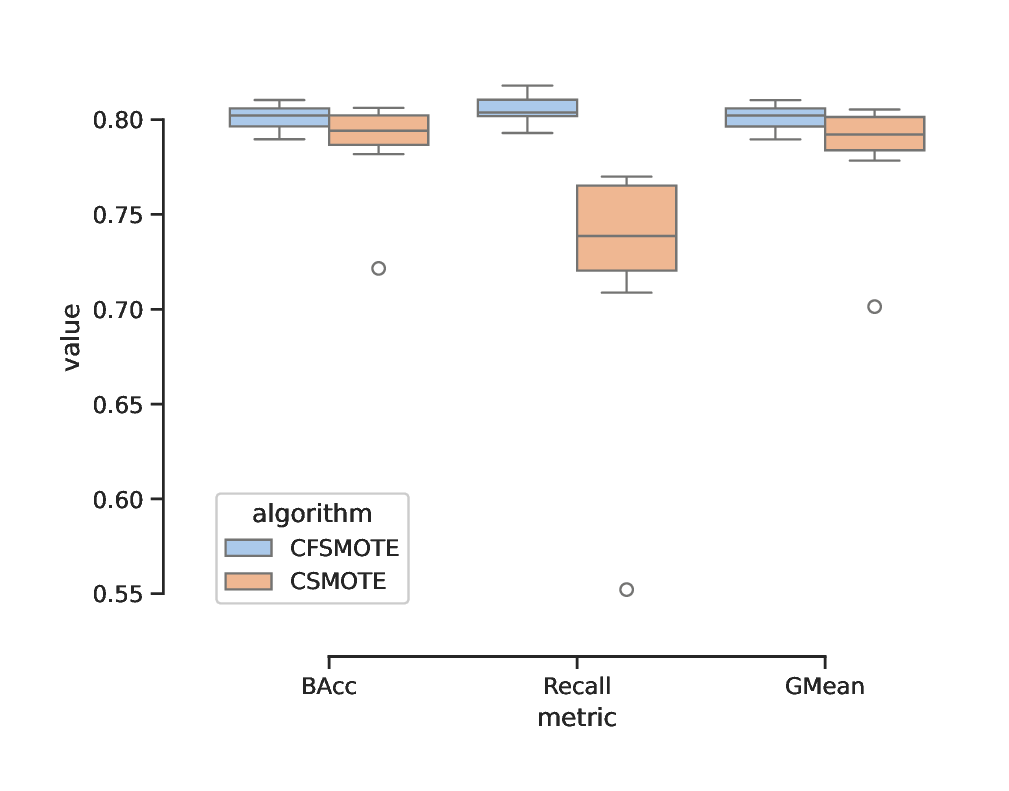}
         \caption{KDD Performance}
     \end{subfigure}%
      \begin{subfigure}[b]{0.5\textwidth}
         \centering      
         \includegraphics[width=\linewidth]{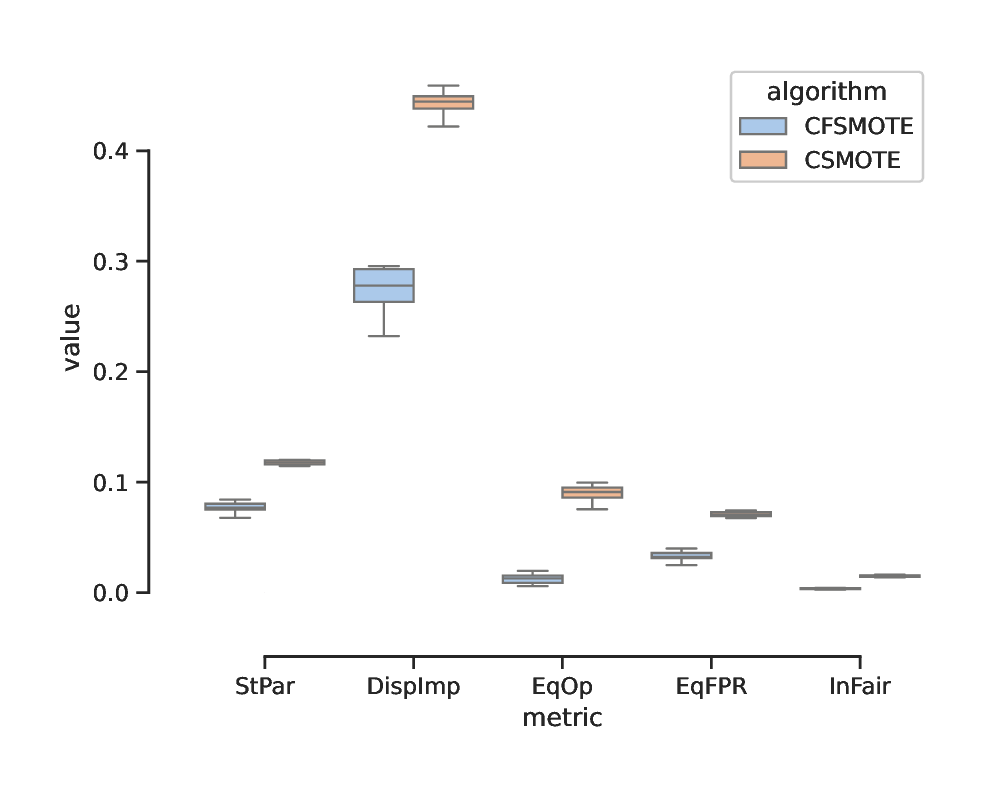}
         \caption{KDD Fairness, smaller better}
     \end{subfigure}
          \begin{subfigure}[b]{0.5\textwidth}
         \centering      
       \includegraphics[width=\linewidth]{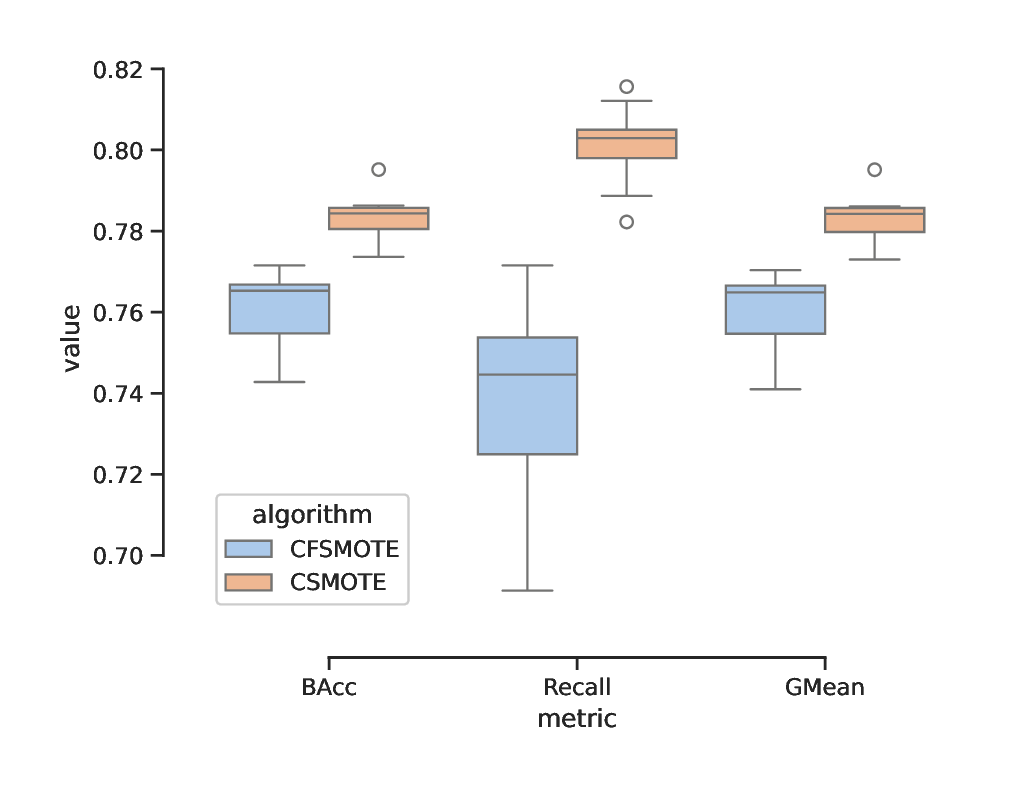}
         \caption{Adult Performance}
     \end{subfigure}%
      \begin{subfigure}[b]{0.5\textwidth}
         \centering      
         \includegraphics[width=\linewidth]{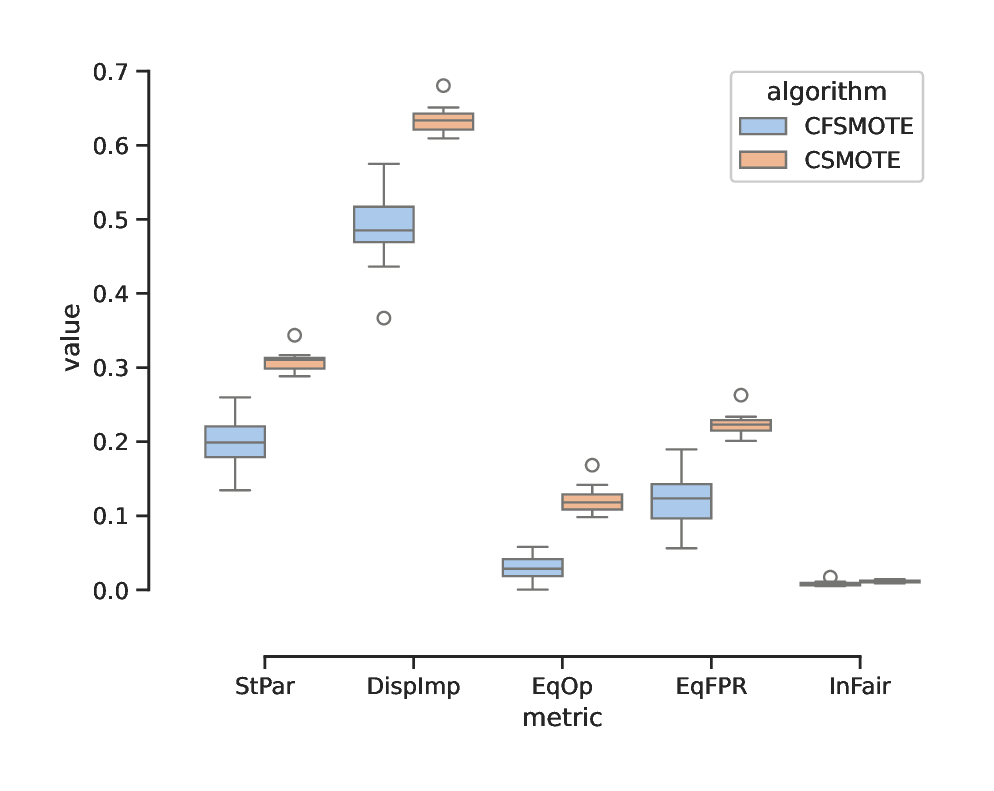}
         \caption{Adult Fairness, smaller better}
     \end{subfigure}
   
    \caption{CFSMOTE vs C-SMOTE}
    \label{fig:cfsmote_vs_csmote}
  \end{figure}
  
The results of our experiments evaluating CFSMOTE in comparison to C-SMOTE are summarized in \cref{tab:cf-vs-csmote} and visualized in \cref{fig:cfsmote_vs_csmote}. 
While we observe a consistent fairness-accuracy-trade-off on the Adult dataset, the performance results on the KDD dataset did not show this effect, with CFSMOTE outperforming C-SMOTE in terms of balanced accuracy, recall and geometric mean. This is likely because CFSMOTE directly optimizes fairness through unawareness and balancing out ratios between all combinations of sensitive attribute and class, i.e. equalized odds. However, even in datasets with observable group disparities in outcomes, optimizing these fairness metrics might not create a significant performance trade-off as long as the base classifier is complex enough to learn all groups sufficiently well, which seems to be the case with the KDD dataset.

On both datasets, CFSMOTE considerably improves on all metrics of fairness, particularly those referred to as equalized odds (equal FPR and equal opportunity). A notable improvement in both disparate impact and statistical parity was likewise apparent. Here, we can clearly show that while C-SMOTE tends to have discrimination scores well above the level of the data stream itself, CFSMOTE achieves results that are well in keeping with the dataset-inherent disparity between the two groups. As CFSMOTE only explicitly considers fairness through unawareness in its implementation, this result was to be expected - while C-SMOTE injects additional unfairness into the algorithm, CFSMOTE simply preserves disparities within the data. In cases where group disparities are justified - for example, through resolving variables - this behaviour is likely what one would desire of a fair classifier. Additionally, we can also observe that the individual fairness is relatively low for both classifiers, which indicates that the base classifier - a decision tree - rarely splits based on the sensitive attribute itself. Therefore, the considerably lower scores on the other discrimination metrics indicate that CFSMOTE also effectively deals with proxy discrimination.

In summary, our first experiment showed that CFSMOTE is capable of considerably improving upon the fairness of C-SMOTE, while a moderate, negative impact on predictive performance could only clearly be observed on one dataset. We, therefore, conclude that our model is a useful alternative for C-SMOTE in contexts where fairness is relevant, particularly in contexts where the data stream itself exhibits justified disparity of outcomes between groups. We could also show that C-SMOTE is prone to significantly exacerbating issues of unfairness and should, therefore, not be the first choice in fairness-relevant situations.

\subsection{CFSMOTE vs Fairness-Aware Stream Learners}
\begin{table}[t]
\caption{Results for the KDD Census Dataset.}
\label{tab:kdd_census}
\centering
\scriptsize

\setlength{\tabcolsep}{0.4em}
\begin{tabular}{l|cccc}
\toprule
 & \multicolumn{1}{c}{CFSMOTE} & \multicolumn{1}{c}{FABBOO
EQOP} & \multicolumn{1}{c}{FABBOO
SP} & \multicolumn{1}{c}{FAHT} \\
\midrule
Balanced Accuracy & $\mathbf{83.01 \pm 0.19}$ & $\underline{80.67 \pm 0.27}$ & $80.27 \pm 0.25$ & $59.46 \pm 0.41$ \\
Recall & $\mathbf{79.69 \pm 1.39}$ & $69.26 \pm 0.60$ & $\underline{71.53 \pm 0.55}$ & $19.73 \pm 0.86$ \\
Geometric Mean & $\mathbf{82.93 \pm 0.24}$  & $\underline{79.86 \pm 0.31}$ & $79.79 \pm 0.28$ & $44.23 \pm 0.94$ \\
\midrule
Statistical Parity & $\:\;9.06 \pm 0.14$ & $\:\;5.80 \pm 0.43$ & $\:\;\mathbf{0.06 \pm 0.03}$ & $\:\;\underline{2.36 \pm 0.11}$ \\
Disparate Impact & $40.37 \pm 2.29$  & $\underline{39.34 \pm 2.96}$ & $\:\;\mathbf{0.38 \pm 0.23}$ & $73.50 \pm 1.44$ \\
Equal Opportunity & $\:\;\underline{4.77 \pm 0.94}$ & $\:\;\mathbf{1.83 \pm 0.91}$ & $\:\;8.77 \pm 1.14$ & $\:\;8.44 \pm 0.87$ \\
Equal FPR & $\:\;4.10 \pm 0.15$ & $\:\;\underline{1.12 \pm 0.41}$ & $\:\;4.56 \pm 0.07$ & $\:\;\mathbf{0.61 \pm 0.05}$ \\
Individual Fairness & $\:\;\underline{1.70 \pm 0.38}$  & $\:\;4.27 \pm 0.48$ & $11.27 \pm 0.34$ & $\:\;\mathbf{0.17 \pm 0.04}$ \\
\bottomrule
\end{tabular}%

\end{table}

\begin{table}[t]

\caption{Results for the Adult Dataset.}
\label{tab:adult_adult}
\centering
\scriptsize

\setlength{\tabcolsep}{0.4em}
\begin{tabular}{l|cccc}
\toprule
 & \multicolumn{1}{c}{CFSMOTE} & \multicolumn{1}{c}{FABBOO
EQOP} & \multicolumn{1}{c}{FABBOO
SP} & \multicolumn{1}{c}{FAHT} \\
\midrule
Balanced Accuracy & $\mathbf{79.23 \pm 0.17}$  & $\underline{78.75 \pm 0.29}$ & $76.81 \pm 0.16$ & $70.61 \pm 0.88$ \\
Recall & $\mathbf{75.53 \pm 0.46}$  & $70.45 \pm 1.17$ & $\underline{73.50 \pm 1.07}$ & $47.58 \pm 2.02$ \\
Geometric Mean & $\mathbf{79.15 \pm 0.19}$ & $\underline{78.31 \pm 0.38}$ & $76.73 \pm 0.19$ & $66.73 \pm 1.30$ \\
\midrule
Statistical Parity & $20.81 \pm 0.53$ & $18.32 \pm 0.46$ & $\:\;\mathbf{0.19 \pm 0.11}$ & $\underline{17.23 \pm 1.41}$ \\
Disparate Impact & $\underline{54.82 \pm 1.16}$ &  $55.89 \pm 1.11$ & $\:\;\mathbf{0.58 \pm 0.33}$ & $78.44 \pm 2.38$ \\
Equal Opportunity & $\:\;\underline{2.30 \pm 1.05}$ & $\:\;\mathbf{1.78 \pm 0.85}$ & $18.41 \pm 0.97$ & $21.97 \pm 3.17$ \\
Equal FPR & $11.20 \pm 0.51$ &  $\:\;\underline{8.45 \pm 0.38}$ & $\:\;9.43 \pm 0.27$ & $\:\;\mathbf{7.57 \pm 1.15}$ \\
Individual Fairness & $\:\;\underline{3.28 \pm 0.70}$  & \:\;$8.50 \pm 1.48$ & $30.34 \pm 0.47$ & $\:\;\mathbf{1.27 \pm 0.82}$ \\
\bottomrule
\end{tabular}%
\end{table}

\begin{figure}[t]
\centering
\centering
     \begin{subfigure}[b]{0.5\textwidth}
         \centering      
         \includegraphics[width=\linewidth]{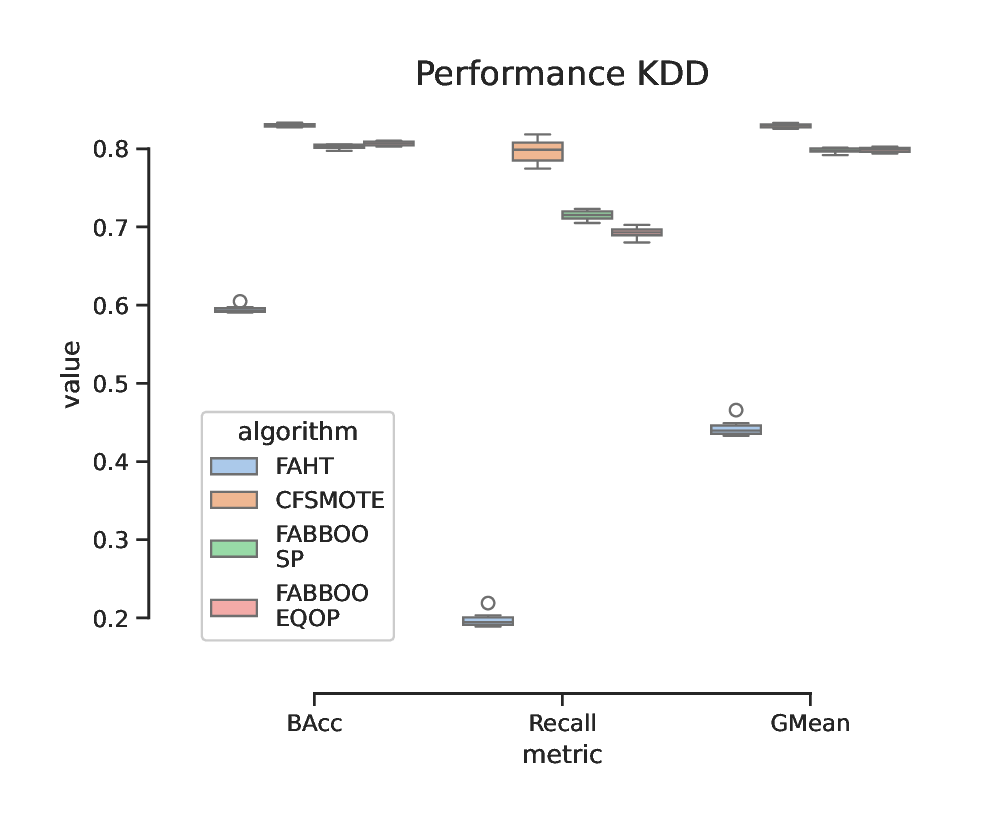}
         \caption{KDD Performance}
     \end{subfigure}%
      \begin{subfigure}[b]{0.5\textwidth}
         \centering      
         \includegraphics[width=\linewidth]{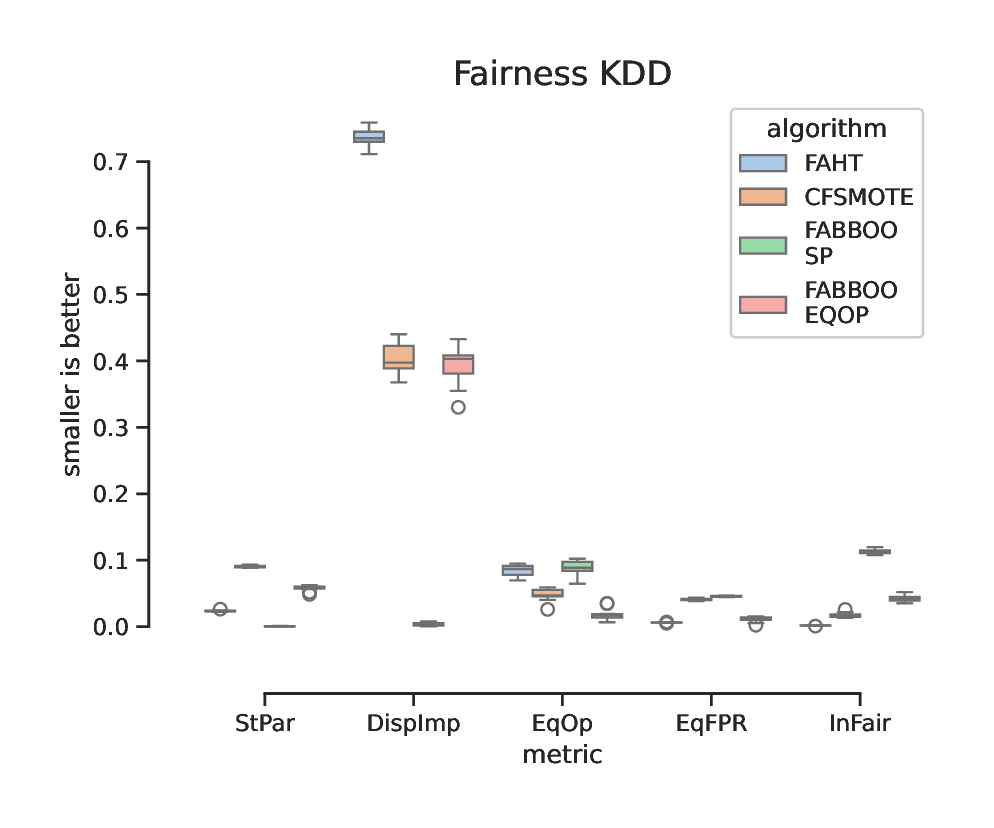}
         \caption{KDD Fairness}
     \end{subfigure}
          \begin{subfigure}[b]{0.5\textwidth}
         \centering      
       \includegraphics[width=\linewidth]{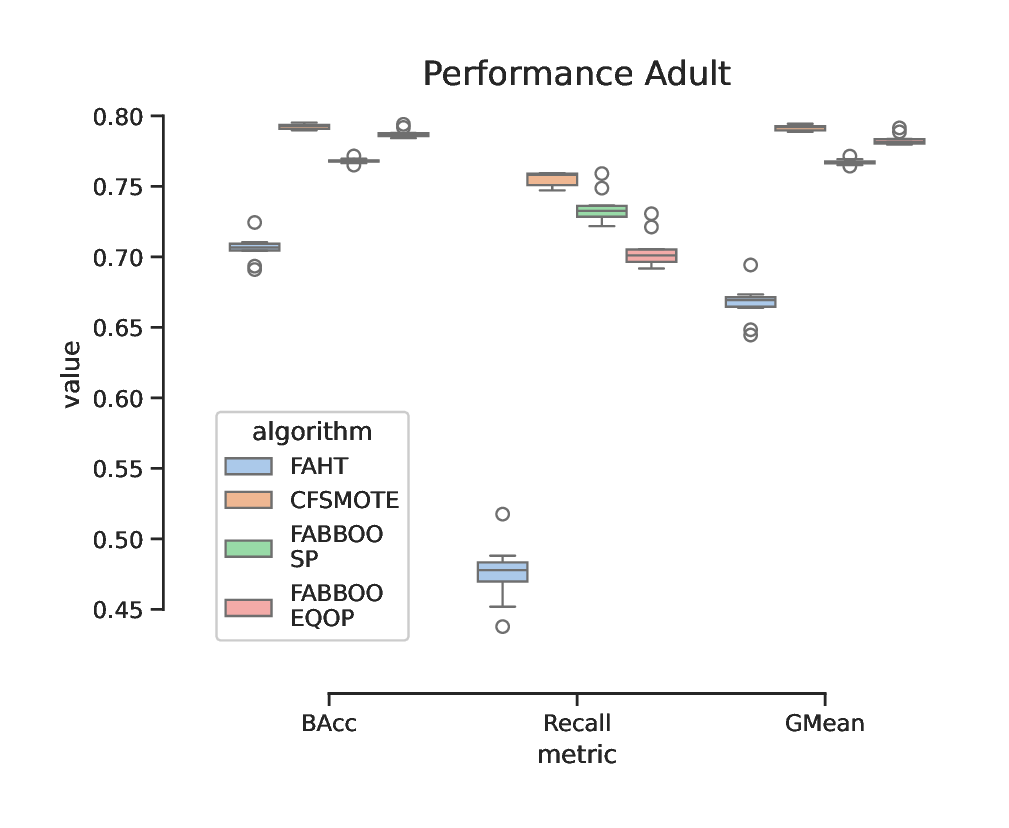}
         \caption{Adult Performance}
     \end{subfigure}%
      \begin{subfigure}[b]{0.5\textwidth}
         \centering      
         \includegraphics[width=\linewidth]{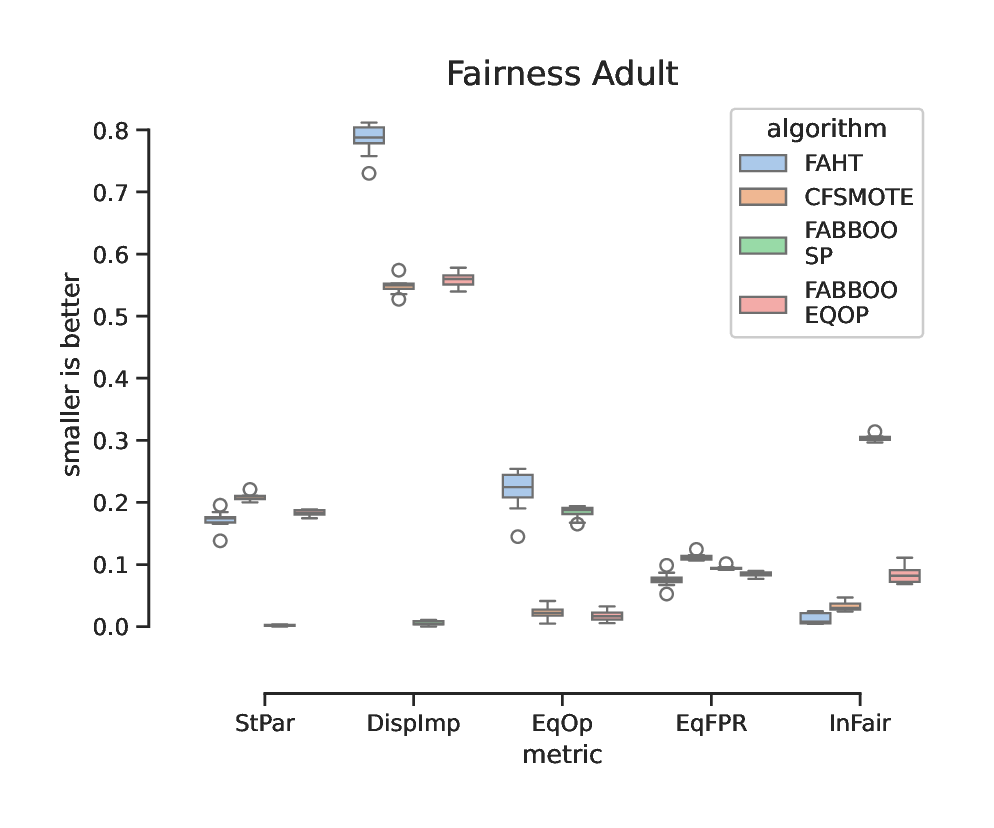}
         \caption{Adult Fairness}
     \end{subfigure}
   
    \caption{Comparison of Fair Learners}
    \label{fig:fair_learner_comparison}
  \end{figure}

Our experiments comparing CFSMOTE with other available fair stream learning methods are summarized out \cref{tab:adult_adult,tab:kdd_census,fig:fair_learner_comparison}. They show that CFSMOTE, in this case with an ARF as a base learner, outperforms the other algorithms in terms of balanced accuracy, recall, and geometric mean. Particularly on the KDD dataset, which exhibits a comparatively strong class imbalance, its performance on the minority class is significantly better than that of the imbalance-aware FABBOO. As expected, FAHT, which is not designed to deal with class imbalance, shows significantly lower performative results than the other classifiers. With FABBOO, relatively high predictive performance was achieved for the version optimizing equal opportunity, while a more notable fairness-accuracy-trade-off was observable for that minimizing statistical parity. This effect was much stronger on the Adult dataset than on the KDD Census data, likely due to the significantly higher inherent disparity of the former.

When analyzing the fairness scores of the models, it becomes clear that FABBOO's post-processing approach is highly effective at minimizing statistical parity, which in turn also leads to very low disparate impact scores. However, this comes with costs to equalized odds and individual fairness, with significant levels of discrimination being detected when judging by those metrics. This shows a clear trade-off between different definitions of fairness, and minimizing one of them will often result in significant increases in the other scores. The FABBOO variant optimizing for equal opportunity also creates similar false-positive rates due to the balanced approach. Interestingly, the post-processing is notably less effective for equal opportunity than for statistical parity, with the latter variant notably closer to the tolerated level of 0.1\% of the optimized discrimination metric. Our experiments also showed that CFSMOTE, in addition to low levels of discrimination on an individual level, also achieves relatively good scores when it comes to equalized odds.

Interestingly, FAHT shows very low levels of discrimination both on the individual fairness metric as well as on equal false-positive rates, though this might be at least partially explained by the fact that the fairness-aware split metric likely avoids splits on the sensitive attribute. The comparatively low Recall scores also indicate that FAHT primarily learns the negative majority class on both Adult and especially the strongly imbalanced KDD dataset, which likely explains the good Equal FPR scores - false positives are simply rare for both groups when applying FAHT to imbalanced data. False negatives, as indicated by the equal opportunity score, are meanwhile produced frequently. But despite the good results on several fairness metrics, the low predictive performance on the minority class makes FAHT ultimately unsuitable for imbalanced data.

For CFSMOTE, in contrast, we can observe low levels of discrimination on the individual level despite the strong predictive performance of the classifier. On the Adult dataset, the equal opportunity score achieved by CFSMOTE is very close to that achieved by FABBOO EQOP. In terms of disparate impact, these two classifiers likewise receive 
very similar scores on both datasets.  However, on imbalanced data, disparate impact and statistical parity can lead to very different evaluations of equality of outcome. The statistical parity score is more strongly affected by the frequency of positive predictions, and CFSMOTE has a noticeably higher recall on both datasets. This indicates that CFSMOTE does not treat the groups more differently FABBOO EQOP but gives out positive classifications overall more frequently. This leads to higher statistical parity scores for CFSMOTE, as this metric computes absolute rather than relative differences.

\section{Conclusion \label{sec:conclusion}}
In this paper, we have proposed CFSMOTE, a fairness-ware, continuous SMOTE variant. Our experiments have shown that CFSMOTE markedly improved several relevant fairness metrics in comparison to the vanilla C-SMOTE algorithm, while a moderate fairness-performance-trade-off could only be seen on one dataset. We have also demonstrated that CFSMOTE achieves competitive scores - both in terms of predictive performance as well as individual fairness, and also several group fairness metrics - in comparison to other fairness-aware stream learners for imbalanced data, particularly FABBOO as optimized for equal opportunity.

However, our research has also shown that there is currently a lack of relevant datasets to serve as benchmarks when evaluating fairness in the stream learning context, containing both drift as well as meaningful features. This should be addressed in future work to allow for a more in-depth analysis of the fairness and performance of stream-learning classifiers beyond group fairness measures.

\bibliographystyle{unsrt}
\bibliography{bibliography}
\end{document}